# NEUROPSYCHIATRIC DEVIATIONS FROM NORMATIVE PROFILES: AN MRI-DERIVED MARKER FOR EARLY ALZHEIMER DISEASE DETECTION

*Synne Hjertager Osenbroch*[1,‡]   *Lisa Ramona Rosvold*[1,‡]
Yao Lu[1,2]   Alvaro Fernandez-Quilez[1,3,†]
*for the Alzheimer's Disease Neuroimaging Initiative*

[1]Department of Computer Science and Electrical Engineering, University of Stavanger, Norway.
[2]Department of Computer Science, University of Putra Malaysia, Selangor, Malaysia.
[3]SMIL, Department of Radiology, Stavanger University Hospital (SUS), Norway.

**ABSTRACT**

Neuropsychiatric symptoms (NPS) such as depression and apathy are common in Alzheimer's disease (AD) and often precede cognitive decline. NPS assessments hold promise as early detection markers due to their correlation with disease progression and their non-invasive nature. Yet current tools cannot distinguish whether NPS are part of aging or early signs of AD, limiting their utility. We present a deep learning–based normative modelling framework to identify atypical NPS burden from structural MRI. A 3D convolutional neural network was trained on cognitively stable participants from the Alzheimer's Disease Neuroimaging Initiative, learning the mapping between brain anatomy and Neuropsychiatric Inventory Questionnaire (NPIQ) scores. Deviations between predicted and observed scores defined the Divergence from NPIQ scores (DNPI). Higher DNPI was associated with future AD conversion (adjusted OR=2.5; $p < 0.01$) and achieved predictive accuracy comparable to cerebrospinal fluid A$\beta$42 (AUC=0.74 vs 0.75). Our approach supports scalable, non-invasive strategies for early AD detection.

## 1. INTRODUCTION

Alzheimer's disease (AD) is the leading cause of dementia, affecting over 55 million people worldwide and contributing to nearly 10 million new cases each year [1]. AD is a progressive and currently non-reversible neurodegenerative disorder. Although no curative treatment exists, early detection can enable interventions such as lifestyle modifications that may delay symptom onset and improve quality of life [2]. In current clinical practice, however, early detection relies on invasive, non-accessible or costly biomarkers [1]. Cerebrospinal fluid (CSF) analysis is widely accepted for identifying underlying AD pathology, particularly A$\beta$ deposition [1]. Its analysis is often combined with cognitive screening tools such as the Mini-Mental State Examination (MMSE) or the Clinical Dementia Rating (CDR) scale [3]. Within this framework, A$\beta$-positive individuals who already show mild cognitive impairment are typically classified as having mild cognitive impairment (MCI), a prodromal stage of AD with high likelihood of progression to dementia [4].

Neuropsychiatric symptoms (NPS) including depression, anxiety, agitation, delusions, apathy, and irritability are highly prevalent in AD [4, 5]. They contribute to accelerated disease progression, reduced quality of life, and increased caregiver burden [6]. NPS are often among the earliest observable changes noticed by families and clinicians, making them promising candidates for early AD detection [6]. Their assessment is scalable, non-invasive and typically carried out using accessible instruments such as the Neuropsychiatric Inventory Questionnaire (NPIQ) [7]. However, while widely used, the NPIQ captures only caregiver-reported symptom burden and does not clarify whether NPS reflect secondary consequences of neurodegeneration, early indicators of AD pathology, or causal risk factors for AD progression [5]. This ambiguity has prevented their systematic integration into early-detection strategies, with a central challenge being how to disentangle NPS profiles from other factors at the earliest disease stages [5, 8].

Recent advances in artificial intelligence (AI) have enabled AD risk stratification by leveraging cognitive test scores, genetic risk factors, or neuroimaging markers such as hippocampal atrophy [3, 9]. Several studies have also incorporated NPS into predictive models together with imaging or biomarker data, but these approaches leave their independent contribution to AD risk unresolved [6]. Normative modelling has emerged as a powerful AI framework to characterize atypicality by comparing individuals against expected population patterns [10]. By training on normative cohorts, deviations from predicted values provide personalized indicators of factors that fall outside a "normal" range [10, 11, 12]. Structural MRI (sMRI) is particularly compelling for this approach: it is non-invasive, widely available, and sensitive to subtle neuroanatomical changes that may underlie NPS [2]. Yet normative modelling has not been systematically applied to neuropsychiatric profiles in AD, and the connection between NPS and MRI remains unexplored.

‡ Equal contribution
† Corresponding author: alvaro.f.quilez@uis.no

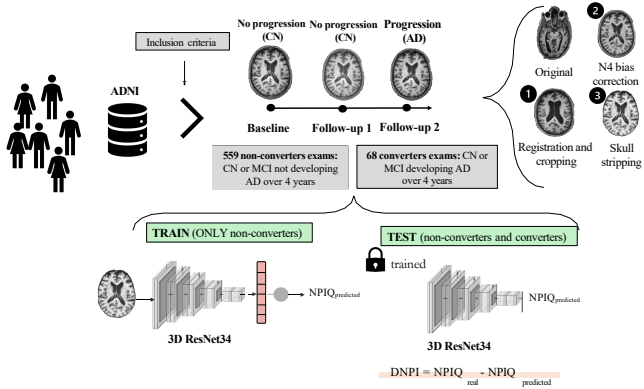

**Fig. 1**: Overview of the study design. A total of 627 brain T1-weighted MRI exams with available CDR, NPIQ, age, MMSE, gender and A$\beta$42 were included. A 3D ResNet34 model was trained to predict NPIQ from the T1w exam of non-converters. DNPI was defined as the residual between the real NPIQ and predicted NPIQ, and tested for association with AD development risk and discriminative abilities for AD development.

In this study, we hypothesize that neuropsychiatric deviations, when quantified through MRI-based normative modelling, serve as early and independent markers of AD risk rather than consequences of the pathology. To test this hypothesis, we defined a normative reference group as cognitively stable individuals with normal values of A$\beta$42, MMSE, and CDR [4] and trained a deep learning framework using their structural MRI to predict NPIQ score. Deviations between predicted and observed scores defined the Divergence from NPI scores (DNPI), capturing abnormal neuropsychiatric profiles relative to normative brain anatomy anchored in biomarker and cognitively-normal populations. We then evaluated DNPI as a scalable, non-invasive marker of early AD risk, benchmarking its prognostic utility against established biomarkers such as amyloid status (A$\beta$42) collected from CSF.

## 2. MATERIALS AND METHODS

### 2.1. Participants and study design

The data used in this study was sourced from Alzheimer's Disease Neuroimaging Initiative (ADNI; adni.loni.usc.edu), a large, ongoing longitudinal study launched in 2003 to understand the progression of AD from normal aging to MCI and early AD [2]. Participants in ADNI span across the diagnostic categories cognitively normal (CN), MCI, and AD [2].

For the purpose of this study, we applied the following inclusion criteria: (i) individuals who had a baseline 1.5 Tesla (T) or 3.0T T1-weighted (T1w) MRI scan and no dementia-unrelated neurological or psychiatric disorders, (ii) diagnosed as either MCI or CN at the first available visit (baseline) (iii) at least two available visits (iv) MMSE total score, clinical diagnosis, age, gender, CDR and NPIQ available for the first visit and (v) baseline MRI scans of diagnostic quality [2, 3].

After applying the inclusion criteria, all eligible participants were divided into two groups: individuals who remained CN or MCI throughout follow-up defined as non-converters, and those who progressed to AD at any subsequent visit within four years of the first available visit (converters). Progression was defined according to the clinical diagnostic classifications available in the ADNI database [2]. A total of 212 individuals with 627 available visits met all inclusion criteria. The final cohort is illustrated in Figure 1.

### 2.2. MRI pre-processing

T1w MRI images were acquired from various commercial scanners including Siemens Healthineers (Erlangen, Germany), Philips Medical Systems (Eindhoven, The Netherlands) and GE Healthcare, (Chicago, Illinois, USA). All images had a native field of view (FOV) of 240x256x208 mm$^3$ and an isotropic resolution of 1x1x1 mm$^3$ [3]. To account for scanner differences, each scan was non-linearly registered to the MNI152 standard space using Advanced Normalization Tools (ANTs), ensuring spatial alignment across subjects for further analysis [2]. Following, all MRI scans were resampled to an isotropic voxel size of 1x1x1 mm and aligned to a resolution of 182x218x182 voxels [3]. After spatial normalization, N4 bias field correction was applied using ANTs to mitigate scanner-related intensity inhomogeneities [2]. Finally, all exams were skull stripped using the Brain Extraction Tool (BET). A fractional intensity threshold of 0.5 was used to control the brain boundary estimation [2] (Figure 1).

### 2.3. Normative psychiatric modelling framework

Normative models were trained to capture the healthy range of NPIQ scores from sMRI, using exclusively non-converter and biomarker-normal group individuals as the normative reference population (Figure 1). This strategy ensured that model predictions reflected neuropsychiatric variability within bio- marker and cognition-normal individuals, enabling deviations at test time to be interpreted as potential indicators of neuropsychiatric atypicality in the context of AD [1].

We adapted a 3D ResNet34 architecture for regression (Figure 1), replacing the classification output layer with a single linear neuron predicting continuous NPIQ scores [10]. Models were trained with the Adam optimizer (learning rate = 0.001), mean squared error loss, and 300 epochs, with checkpointing based on validation loss. To improve generalization, dynamic data augmentation was applied at training time, including random 90° rotations, random flips along each axis, additive Gaussian noise, and random intensity scaling and shifting [3, 12, 13].

Data were split into disjoint training, validation, and test sets to avoid subject overlap. The training set comprised 62% of non-converter exams (N=345). The validation set included 25% of non-converter exams (N=141) and 25% of converter exams (N=14), while the test set consisted of 13% of non-converter exams (N=73) and 75% converter exams (N=54).

### 2.4. Divergence from NPI

During testing, DNPI was defined as the residual obtained by subtracting the observed NPIQ score from the score predicted by the MRI-based normative framework (Figure 1). DNPI thus quantifies the degree to which a subject's neuropsychiatric profile deviates from what would be expected in a biomarker and cognition-normal reference population. Under this assumption, the model captures thus normative brain–behavior patterns from cognitively stable individuals, and large DNPI values in test subjects indicate neuropsychiatric manifestations not aligned with normal variation, potentially reflecting early pathological processes in the context of AD [1].

### 2.5. Statistical analysis

All analyses were performed in Python 3.10 (https://www.python.org) using the open-source statsmodels 0.14.0 package (https://www.statsmodels.org/stable/index.html). Continuous variables were summarized as mean ± standard deviation, while categorical variables were reported as counts and percentages (N [%]). Group differences were assessed using unpaired t-tests or Mann–Whitney U tests, depending on data distribution.

Logistic regression was employed for two purposes: (i) association testing and (ii) discriminative modeling [10]. For association testing, logistic regression was used to determine whether DNPI was associated with conversion to AD. DNPI was introduced as a continuous predictor, and both unadjusted and adjusted models were fitted. Covariates included age, gender, APOE4 status, CDR, MMSE, and amyloid status. Results were reported as odds ratios (OR) with 95% confidence intervals (CI) and p-values, where the OR represents the change in odds of conversion associated with a one-unit increase in DNPI.

For discriminative modeling, logistic regression models were trained on the validation set and applied to the independent test set. Two clinically relevant scenarios were evaluated: (a) univariate models using DNPI or $A\beta42$ as single predictors, and (b) multivariate models integrating DNPI or $A\beta42$ with age, gender, and CDR, reflecting typical diagnostic pathways.

Predictive performance was assessed using area under the ROC curve (AUC), balanced accuracy (BA), sensitivity, specificity and F1 scores [3, 10]. To reflect clinical priorities, confusion matrices were reported at a fixed false positive rate (FPR) of 0.20 [4]. AUC confidence intervals were estimated using 1,000 bootstrap iterations, and confusion matrices were generated to provide additional insight into classification performance. A p-value<0.05 was considered statistically significant.

**Table 1**: Association between a one-unit increase in DNPI and AD conversion across adjustment models. Odds ratios (OR) with 95% confidence intervals (CI) and p-values are reported.

| Model | OR | 95% CI | p-value |
|---|---|---|---|
| DNPI (Unadjusted) | 2.50 | [1.40, 4.47] | 0.002† |
| + Gender | 2.50 | [1.40, 4.47] | 0.002† |
| + Age | 2.61 | [1.44, 4.70] | 0.001† |
| + CDR | 2.60 | [1.43, 4.70] | 0.002† |
| + MMSE | 2.45 | [1.32, 4.56] | 0.005† |
| + $A\beta42$ | 2.25 | [1.25, 4.02] | 0.007† |
| + Gender, Age, CDR, $A\beta42$ | 2.70 | [1.48, 4.90] | 0.001† |

DNPI = Divergence from NPI, CDR = Clinical Dementia Rating, MMSE = Mini-Mental State Examination, $A\beta42$ = CSF amyloid-beta deposition
†Statistically significant (p-value<0.05).

## 3. RESULTS

The study included 627 visits from 212 participants from the ADNI cohorts (Figure 1). Converters were on average older (75.69±6.87 vs 72.57±7.47), showed lower MMSE (25.45±2.8 vs 28.41±1.33), higher CDR scores (0.50 ± 0.12 vs 0.24±0.26) and had a higher male prevalence (63.6% vs 52.9%) compared to non-converters.

Logistic regression analyses showed that DNPI was significantly associated with conversion to AD across all adjustment models. As summarized in Table 1, each one-point increase in DNPI was linked to a 2.25–2.70 fold increase in the odds of conversion, with all associations remaining significant ($p < 0.01$) after adjusting for $A\beta42$, CDR, MMSE, age, and gender.

In predictive analyses, DNPI outperformed univariate $A\beta42$ models and demonstrated competitive accuracy when integrated with clinical covariates. We benchmarked DNPI against $A\beta42$ in clinically relevant diagnostic scenarios (Table 2): (a) DNPI achieved higher balanced accuracy (0.65 vs. 0.54) and sensitivity (0.70 vs. 0.44) compared to $A\beta42$, translating into fewer false negatives while maintaining comparable specificity and (b) when integrated with age, gender, and CDR, DNPI retained superior sensitivity (0.72 vs. 0.63) and F1 score (0.64 vs. 0.58), despite A$\beta$42 with covariates achieving a slightly higher AUC (0.75 vs. 0.74).

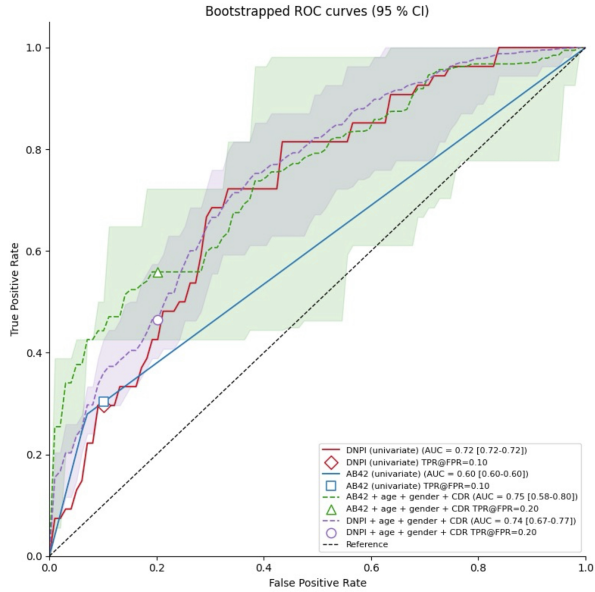

**Fig. 2**: Bootstrapped ROC curves (95% CI) for DNPI and benchmarks. Shaded areas highlight bootstrap intervals around the specific AUCs

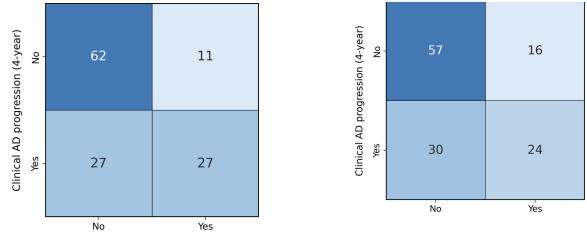

(a) DNPI @ FPR = 0.20   (b) A$\beta$42 @ FPR = 0.20

**Fig. 3**: Confusion matrices on the independent test set at a fixed false positive rate (FPR) of 0.20.

Bootstrapped ROC analyses confirmed the discriminative advantage of DNPI (AUC = 0.72 [95% CI: 0.70–0.74]) over A$\beta$42 (AUC = 0.60 [95% CI: 0.58–0.62]) (Figure 2). When integrated with covariates, both DNPI and A$\beta$42 achieved higher AUCs (0.74 and 0.75, respectively), though DNPI retained superior sensitivity. Confusion matrices at the clinical operating point (FPR = 0.20) illustrate the trade-offs in classification (Figure 3). DNPI correctly identified a higher number of converters compared to A$\beta$42, reducing false negatives while maintaining comparable specificity.

**Table 2**: Predictive performance of DNPI and benchmark models on the independent test set with 95% confidence intervals (CI).

| Model | BA | AUC | F1 |
|---|---|---|---|
| DNPI (univariate) | 0.65 [0.63-0.67]† | 0.72 [0.70–0.74]† | 0.62[0.60-0.64]† |
| A$\beta$42 (univariate) | 0.54 [0.53-0.58] | 0.60 [0.59-0.61] | 0.45 [0.43–0.47] |
| A$\beta$42 + age + gender + CDR | 0.62 [0.60–0.65] | 0.75 [0.59–0.80] | 0.58 [0.56–0.60] |
| DNPI + age + gender + CDR | 0.66 [0.64–0.68]† | 0.74 [0.67–0.77]† | 0.64 [0.62–0.66]† |

DNPI = Divergence from NPI, CDR = Clinical Dementia Rating, MMSE = Mini-Mental State Examination, BA = Balanced Accuracy
†Statistically significant (p-value<0.05), compared against A$\beta$42 model.

## 4. DISCUSSION

This study introduces an MRI-based normative modeling framework to quantify neuropsychiatric deviations as early markers of AD. To our knowledge, this is the first application of normative modeling to link structural MRI with neuropsychiatric symptom profiles in AD. By establishing individualized deviations from cognitively stable populations, our approach disentangles psychiatric burden from typical variation, providing a scalable, non-invasive biomarker of AD risk.

Compared to established amyloid-based workflows [1], DNPI demonstrated higher sensitivity and balanced accuracy in univariate analyses, and retained competitive performance when integrated with demographic and clinical covariates. Prior AI models have included NPS as predictors alongside imaging or biomarker data [6, 5], but none have treated them as outcomes inferred from brain structure. Our results show that this perspective yields complementary prognostic value, directly linking brain anatomy to behavioral phenotypes.

Clinically, this approach could support early risk stratification in scenarios where invasive or costly biomarkers are unavailable [3]. MRI is already part of routine clinical practice in aging populations [3], and DNPI can be derived without additional procedures, potentially improving accessibility in early detection of AD. This work has limitations. Our normative models were trained and validated within ADNI, which may limit generalizability to more diverse populations. Further studies are needed to validate DNPI across cohorts and to explore integration with longitudinal trajectories.

## 5. CONCLUSIONS

We presented a deep learning–based normative modeling framework to derive individualized neuropsychiatric profiles from structural MRI. By quantifying deviations from a cognitively stable reference population, we introduced DNPI as a novel imaging-derived marker of early AD risk. Our results indicate that DNPI is strongly associated with AD conversion, improves predictive performance compared to established amyloid-based markers, and can offer a non-invasive alternative to current diagnostic workflows.

## 6. COMPLIANCE WITH ETHICAL STANDARDS

This research leveraged open access ADNI data. Ethical approval was not required, as confirmed by ADNI data license.

## 7. ACKNOWLEDGMENTS